\documentclass{style/naturep}

\usepackage{authblk}
\usepackage{graphicx,booktabs,array,multirow,amsmath,amssymb,xcolor}
\usepackage{ragged2e,setspace,hyperref,caption,geometry}
\makeatletter
\let\saved@includegraphics\includegraphics
\@namedef{nbme@figureheight@figures/out/fig1.pdf}{\dimexpr\textheight-99.0pt\relax}
\@namedef{nbme@figureheight@figures/out/fig2_board2.pdf}{\dimexpr\textheight-99.0pt\relax}
\@namedef{nbme@figureheight@figures/out/fig3_distill.pdf}{\dimexpr\textheight-99.0pt\relax}
\@namedef{nbme@figureheight@figures/out/fig4_adjudication.pdf}{\dimexpr\textheight-142.5pt\relax}
\@namedef{nbme@figureheight@figures/out/fig5_examples.pdf}{\dimexpr\textheight-171.5pt\relax}
\@namedef{nbme@figureheight@figures/out/fig6_efficiency.pdf}{\dimexpr\textheight-171.5pt\relax}
\AtBeginDocument{%
  \renewcommand{\includegraphics}[2][]{%
    \ifcsname nbme@figureheight@#2\endcsname
      \saved@includegraphics[#1,height=\csname nbme@figureheight@#2\endcsname,keepaspectratio]{#2}%
    \else
      \saved@includegraphics[#1]{#2}%
    \fi}%
}
\makeatother

\newcommand{\bmhead}[1]{\section*{#1}}

\begin{document}

\title{\justifying\bfseries NIBench: benchmarking autonomous AI agents for execution, skill evolution and scientific judgment in neuroimaging}

\author[1,2,$^{\ddagger}$]{Lujia Zhong}

\author[1,$^{\ddagger}$]{Yihao Xia}

\author[1,2,$^{\ddagger}$]{Jianwei Zhang}

\author[1,3]{Shuo Huang}

\author[1]{Xinyu Nie}

\author[1,2]{Mingyang Xia}

\author[1,2]{Zhiwei Deng}

\author[1,2]{Jiaxin Yue}

\author[1,2]{Xinkai Wang}

\author[4,5,6]{Junhan Zhao}

\author[1,2,3]{Yonggang Shi}

\affil[1]{{Stevens Neuroimaging and Informatics Institute, Keck School
of Medicine}, {University of Southern California},
{{Los Angeles}, {CA}, {USA}}}

\affil[2]{{Ming Hsieh Department of Electrical and Computer Engineering,
Viterbi School of Engineering}, {University of Southern California},
{{Los Angeles}, {CA}, {USA}}}

\affil[3]{{Alfred E. Mann Department of Biomedical Engineering, Viterbi
School of Engineering}, {University of Southern California},
{{Los Angeles}, {CA}, {USA}}}

\affil[4]{Department of Pediatrics, University of Chicago, Chicago, IL, USA}

\affil[5]{Department of Biomedical Informatics, Harvard Medical School, Boston, MA, USA}

\affil[6]{Comprehensive Cancer Center, UChicago Medicine, Chicago, IL, USA}

\affil[$^{\ddagger}$]{These authors contributed equally to this work.}

\begin{spacing}{1.2}
\maketitle
\par\medskip
\noindent Corresponding author: Yonggang Shi (yshi@loni.usc.edu)

\begin{abstract}
%% Abstract wording approved by the authors on 2026-10-12.

Autonomous AI agents are increasingly used to automate biomedical research, creating a growing need to benchmark their capabilities on real scientific data and workflows. Their reliability in neuroimaging remains insufficiently characterized, where plausible analytical outputs can conceal errors that affect scientific conclusions. Existing studies have advanced scientific question answering and tool-based automation, but provide limited evidence on workflow reconstruction, acquisition of procedural expertise and evidence-based scientific judgment. Here we present NIBench, a neuroimaging benchmark comprising three complementary evaluations of autonomous agents. First, we benchmark execution and workflow reconstruction using 56 tasks across seven tool ecosystems, covering structural, diffusion and functional magnetic resonance imaging, positron emission tomography and multimodal analyses. Second, we examine whether agents can evolve their procedural skills through distillation of execution trajectories, evaluating the resulting skills on previously encountered tasks, unseen tasks and other models. Third, we assess scientific judgment using 36 claims from published Alzheimer’s disease studies, requiring agents to analyse raw cohort data and provide reproducible evidence. We evaluate 19 language models within a shared agent framework, with selected models undergoing skill distillation and claim evaluation. Leading agents execute supplied analysis protocols nearly perfectly, whereas withholding procedural guidance widens performance differences. Distilled skills improve performance on acquisition tasks and transfer between models, but benefits on unseen tasks are inconsistent. Agents reproduce their reported results yet sometimes reach conclusions inconsistent with expert-derived answers, and access to data does not always improve judgment. NIBench provides a systematic reference for evaluating and improving autonomous agents for neuroimaging research, guiding future development across workflow execution, acquisition of procedural expertise and scientific judgment.

\end{abstract}


%% NMI: unheaded introduction, ~700 words, 4 paragraphs.
%% Logic borrowed from the NMI benchmark precedent (s42256-025-01152-1):
%%   agents enter science -> a specific unresolved failure mode -> existing
%%   evaluations cannot see it -> therefore this apparatus -> three questions.
%% NOT "here is our benchmark". The benchmark is the instrument.

Autonomous AI agents increasingly write code, operate scientific software and revise workflows in response to intermediate results \cite{bran2023chemcrow,boiko2023coscientist,swanson2025virtuallab,biomedagent2026}. These abilities create opportunities to automate substantial parts of scientific analysis. Their value for research, however, depends on whether successful task completion translates into scientifically sound analysis and conclusions. Evaluating this requires examining both agents' operational capabilities and the scientific decisions underlying their outputs.

For neuroimaging researchers, autonomous agents offer the prospect of reducing repetitive work and supporting complex analyses across imaging modalities, participant data and specialized software. Researchers therefore need evidence about which analytical responsibilities they can entrust to agents and under what conditions, making domain-specific benchmarks essential for assessing their usefulness. Reliability depends on applying domain knowledge to select appropriate participants, measurements and methods, and to determine what conclusions the resulting evidence supports. The importance of these decisions is well documented: independent teams analysing a shared neuroimaging dataset can reach different conclusions, and differences between preprocessing pipelines can affect brain-wide association findings \cite{botvinik2020variability,li2024variation}. Reproducing an analysis also does not by itself establish the validity or generalizability of its findings \cite{kiar2024variation}. These observations indicate that evaluating agents requires attention to the scientific appropriateness of their decisions as well as the correctness of their outputs. Such evaluation should help researchers select agents and determine how to use them, while identifying the capabilities that developers need to improve.

Adjacent biomedical fields have addressed these evaluation needs through benchmarks tailored to different research and clinical activities. Medical knowledge and clinical-text evaluations, including MultiMedQA, BRIDGE and cross-task medical comparisons, assess answer quality beyond examination accuracy and reveal task-dependent performance \cite{singhal2023clinical,wu2026bridge,tordjman2025deepseek}. Interactive clinical evaluations, including MIMIC-CDM, AgentClinic and MedAgentBench, test information gathering, sequential decision-making and actions in simulated clinical or electronic health record environments, capabilities that static question-answering scores cannot establish \cite{hager2024clinical,schmidgall2026agentclinic,jiang2025medagentbench}. In scientific data analysis, BioDSBench exposes incorrect analyses despite models' code-generation abilities \cite{wang2026biodsbench}, while BioMed-AQA and SciToolEval evaluate experience-assisted task completion and specialized tool coordination \cite{biomedagent2026,ding2025scitoolagent}. Together, these studies show how different evaluation settings reveal distinct aspects of AI capability. Within neurology and brain imaging, multimodal benchmarks assess image interpretation and clinical reasoning through tasks such as lesion localization, differential diagnosis and diagnostic explanation \cite{jing2026neuralmedbench,bercea2025nova,peng2026omnibrainbench}. In neuroimaging, benchmarks accompanying NeuroClaw and NEURA provide important evaluations of workflow quality, tool planning and claim--evidence consistency \cite{wang2026neuroclaw,xie2026neura}. These evaluations provide important evidence, but further characterization is needed to establish what different forms of task success demonstrate about an agent's ability to support scientific analysis. For neuroimaging researchers, completing a supplied procedure, independently developing a workflow, learning from experience and drawing a supported conclusion imply different kinds of competence. An evaluation framework that distinguishes these capabilities can make benchmark performance more informative for research use and future development.

To evaluate agents against these needs, we organize their scientific capabilities into four related dimensions. Execution concerns correctly implementing a specified procedure and producing the required analytical outputs, a foundation for reliable assistance across specialized tools \cite{ding2025scitoolagent,xie2026neura}. Workflow reconstruction concerns independently developing a workflow that meets a stated scientific goal and its constraints. Useful agents should draw on procedural knowledge comparable to that of an experienced researcher, allowing users to specify objectives without prescribing every operational step. The sensitivity of neuroimaging findings to analytical choices makes the quality of these independently developed workflows consequential \cite{botvinik2020variability,li2024variation}. Acquisition of procedural expertise concerns converting experience into reusable knowledge that improves subsequent analyses. Its value depends on whether acquired knowledge reduces repeated instruction and carries over to new tasks and other models, motivating explicit evaluation of transfer \cite{biomedagent2026}. Scientific judgment concerns assessing whether an analysis and its evidence support a scientific claim. This requires considering the suitability of the data, measurements and analytical assumptions, as reproducibility alone cannot establish scientific validity or generalizability \cite{kiar2024variation}. Distinguishing these dimensions helps establish the scope and limits of the scientific competence demonstrated by an agent's performance.

Here we introduce NIBench, a neuroimaging benchmark organized into three complementary evaluations of workflow execution and reconstruction, acquisition and transfer of procedural expertise, and scientific judgment. First, its workflow evaluation comprises 56 tasks across seven tool ecosystems, with each task evaluated under two controlled conditions. We define an analysis protocol as procedural instructions specifying the tools, processing steps and parameter settings used to achieve an analytical objective. In the protocol-guided condition, agents receive this protocol; in the protocol-unspecified condition, it is withheld while the scientific goal, input data and required outputs remain unchanged. The paired conditions distinguish agents' ability to execute supplied procedures from their ability to reconstruct workflows independently. The task suite also tests whether agents recognize data problems that do not trigger software errors. Second, we examine whether agents can acquire reusable procedural knowledge without updating model weights. We distil execution trajectories into editable skill files and test their effects on the tasks used to acquire them, held-out tasks and recipient models. Third, we evaluate scientific judgment using 36 claims from three published Alzheimer’s disease studies. Agents analyse unfiltered Alzheimer’s Disease Neuroimaging Initiative tables and must provide a supported, contradicted or undetermined conclusion, the evidence underlying that conclusion and code that regenerates it. A no-data control tests whether access to data and analytical tools improves conclusions beyond what models can infer from prior knowledge alone \cite{luo2025brainbench}. Together, these evaluations characterize agents' ability to execute procedures, reconstruct workflows, acquire transferable expertise and reach evidence-supported conclusions, while testing how performance depends on procedural guidance, distilled experience and access to data. By identifying where these capabilities succeed and where further improvement is needed, NIBench provides a systematic reference for developing autonomous agents capable of supporting neuroimaging research.

\section{Results}\label{sec:results}

\subsection{The Overview of NIBench}\label{sec:res-bench}\

\begin{figure}[tp]
\centering
\includegraphics[width=\linewidth]{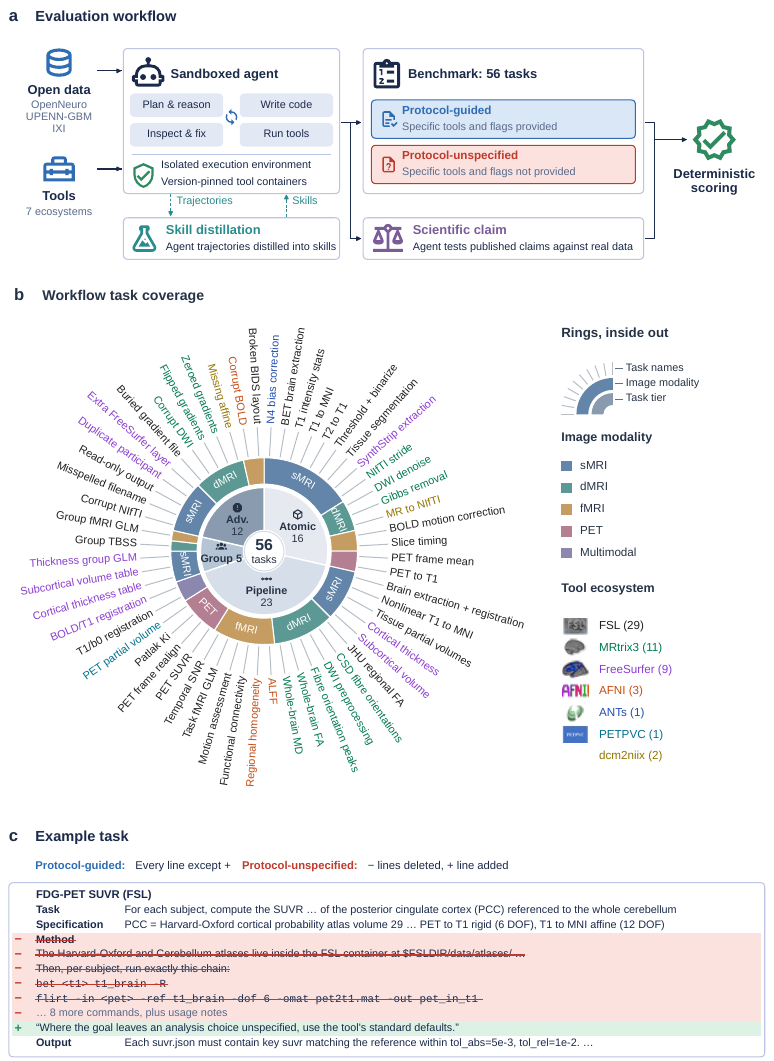}
\caption{\textbf{Overview of NIBench and the workflow evaluation tasks.}
\textbf{a}, The evaluation workflow. \textbf{b}, Distribution of the 56 workflow tasks by task tier, imaging modality and primary tool ecosystem. The inner and middle rings indicate task tier and imaging modality, respectively. Radial labels identify individual tasks, with text colors indicating the primary tool ecosystem.
\textbf{c}, Example of a workflow task under protocol-guided and protocol-unspecified instructions. The protocol-unspecified condition removes the procedural instructions marked with a minus sign and adds the instruction marked with a plus sign. The task objective and output requirements remain unchanged.}\label{fig:decomposition}
\end{figure}

NIBench comprises three evaluation components assessing workflow execution and reconstruction, acquisition and transfer of procedural expertise, and evidence-supported scientific judgment (Fig.~\ref{fig:decomposition}a). Using a shared agent framework, we evaluate 19 language models (11 closed-weight and eight open-weight) on 56 workflow tasks: 16 single-tool operations, 23 multistep pipelines, five group analyses and 12 tasks with problematic inputs, including corrupted images and zeroed diffusion-gradient vectors (Fig.~\ref{fig:decomposition}b). The suite spans structural, diffusion and functional MRI, PET and multimodal analyses across seven tool ecosystems, using data from 35 participants in OpenNeuro ds000114 and ds002898, IXI and UPENN-GBM.

Each workflow task is evaluated with and without an analysis protocol specifying tools, processing steps and parameter settings. The protocol-guided and protocol-unspecified conditions retain identical objectives, inputs and required outputs, distinguishing execution of supplied procedures from independent workflow reconstruction (Fig.~\ref{fig:decomposition}c). Skill evolution is evaluated in five models by distilling execution trajectories from 28 tasks into editable skills without updating model weights, then evaluating performance on these tasks and 28 held-out tasks, together with cross-model skill transfer. Scientific judgment is evaluated in Claude Opus 5.5 and GPT-6-astra using 36 claims from three Alzheimer's disease studies concerning longitudinal cognitive change, tau trajectories and disease progression. The supplied unfiltered Alzheimer's Disease Neuroimaging Initiative tables span 5,029 participants, with imaging-derived measurements linked to more than 19,000 distinct amyloid-PET, tau-PET and structural MRI image identifiers, alongside longitudinal clinical and cognitive assessments, fluid biomarkers and APOE genotypes. Agents independently select participants, measurements and analyses from this data pool for each scientific question; evaluation assesses numerical reproducibility and agreement with reference conclusions.

% \subsection{Withholding an analysis protocol exposes the gap}\label{sec:res-e1}
\subsection{Withholding analysis protocols widens the performance gap between closed- and open-weight models}\label{sec:res-e1}
\begin{figure}[t!]
\centering
\includegraphics[width=\linewidth]{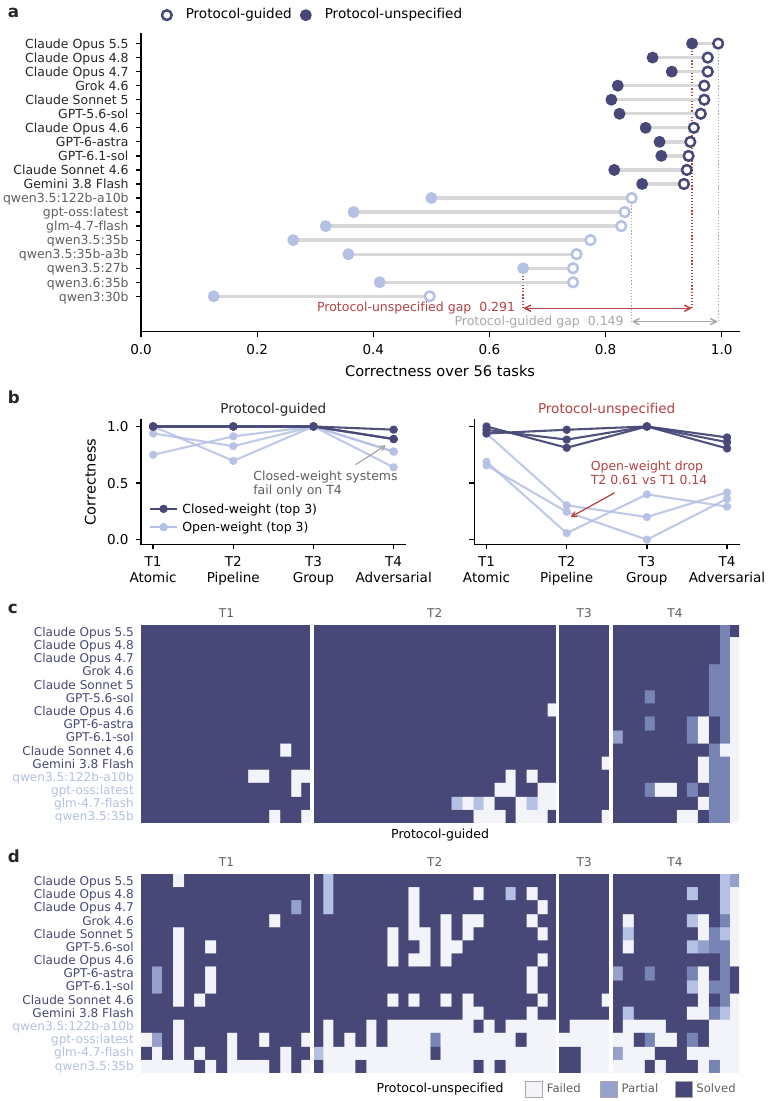}
\caption{\textbf{Execution correctness with and without an analysis protocol.} \textbf{a}, Mean correctness across 56 tasks for 11 closed-weight and eight open-weight models under protocol-guided (open circles) and protocol-unspecified (filled circles) conditions.
\textbf{b}, Correctness by task tier for the three highest-scoring models in each group. \textbf{c},\textbf{d},  Task-level outcomes under protocol-guided (c) and protocol-unspecified (d) conditions for the 11 closed-weight models and four highest-performing open-weight models.}\label{fig:board}
\end{figure}

We compared workflow correctness under protocol-guided and protocol-unspecified instructions across 19 language models using the same agent framework (Fig.~\ref{fig:board}a). Correctness is the mean task score across the 56 tasks, including partial credit. Claude Opus 5.5 achieved the highest correctness in both conditions: 0.994 with the protocol supplied, fully solving 55 of 56 tasks, and 0.949 with the protocol withheld. The highest open-weight scores were 0.845 for qwen3.5:122b-a10b under protocol-guided instructions and 0.658 for qwen3.5:27b under protocol-unspecified instructions. All 11 closed-weight models outscored all eight locally served open-weight models in both conditions. Under protocol-guided instructions, scores ranged from 0.935 to 0.994 for closed-weight models and from 0.497 to 0.845 for open-weight models; under protocol-unspecified instructions, the corresponding ranges were 0.810--0.949 and 0.125--0.658.

Withholding the analysis protocol reduced correctness in all 19 models, with a mean decrease of 0.09 for closed-weight models and 0.38 for open-weight models. Closed-weight models lost at most 0.16, with the smallest decreases of 0.045--0.054 observed for Claude Opus 5.5, GPT-6.1-sol and GPT-6-astra. Open-weight decreases ranged from 0.09 to 0.51. Consequently, the difference between the highest closed-weight and open-weight scores widened from 0.149 under protocol-guided instructions to 0.291 under protocol-unspecified instructions (Fig.~\ref{fig:board}a). This separation emerged while the analytical objectives, input data and required outputs remained unchanged, indicating greater dependence on supplied procedural guidance among the evaluated open-weight models.

The performance loss was particularly pronounced in multistep pipelines (Fig.~\ref{fig:board}b). Among the three highest-scoring open-weight models, mean correctness decreased by 0.61 on pipeline tasks, compared with 0.14 on single-tool operations. The three highest-scoring closed-weight models solved every single-tool, pipeline and group-analysis task under protocol-guided instructions. The task-level results showed a similarly broad effect of withholding procedural guidance: among the 15 models shown in Fig.~\ref{fig:board}c,d, the number of tasks fully solved by every model fell from 29 of 56 to two of 56. Together, these comparisons show that successful execution of a supplied protocol can substantially overestimate performance when agents must independently reconstruct the workflow, especially for analyses requiring multiple processing steps.

\subsection{Skill distillation improves performance on familiar tasks but yields limited generalization}\label{sec:res-sd}
\begin{figure}[t!]
\centering
\includegraphics[width=\linewidth]{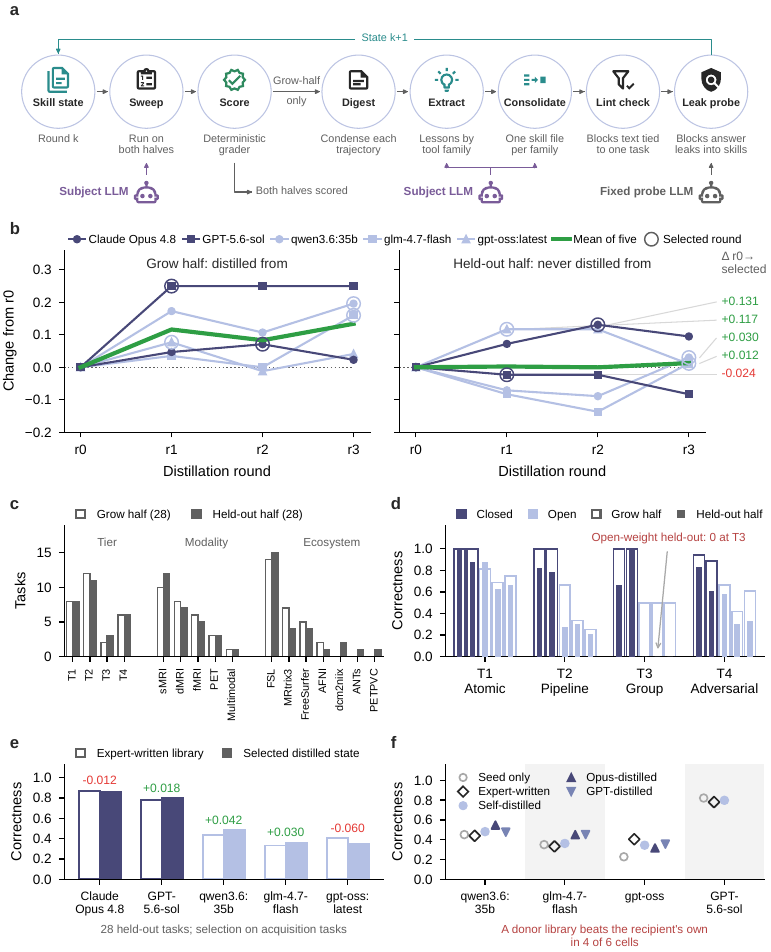}
\caption{\textbf{Acquisition and transfer of procedural skills through trajectory distillation.} \textbf{a}, Skill-distillation procedure. \textbf{b}, Change in correctness relative to round 0 across successive distillation rounds for the 28 acquisition tasks (left) and 28 held-out tasks (right).
\textbf{c}, Distribution of acquisition and held-out tasks by tier, imaging modality, and primary tool ecosystem.
\textbf{d}, Correctness by task tier for closed-weight and open-weight models at their selected distillation rounds.
\textbf{e}, Held-out correctness using the selected distilled library compared with the expert-written library. \textbf{f}, Held-out correctness using different skill libraries, including the initial seed, expert-written, self-distilled, and cross-model distilled libraries.}\label{fig:distill}
\end{figure}

We evaluated skill distillation in five models, using execution trajectories from 28 tasks to construct reusable skill libraries and assessing performance on these acquisition tasks and 28 held-out tasks. The two halves were balanced by task tier, imaging modality and tool ecosystem. For each model, we selected the distilled state with the highest acquisition-task correctness, without using held-out performance for selection (Fig.~\ref{fig:distill}a--c).

Distillation improved acquisition-task correctness in all five models, with gains ranging from 0.071 to 0.250, whereas held-out changes ranged from $-0.024$ to $+0.131$ (Fig.~\ref{fig:distill}b). GPT-5.6-sol showed the largest acquisition gain, increasing from 0.726 to 0.976, while its held-out correctness decreased from 0.821 to 0.798. Qwen3.6:35b and GLM-4.7-flash gained 0.196 and 0.161 on acquisition tasks but only 0.030 and 0.012 on held-out tasks. Transfer to unseen tasks was more evident for Claude Opus 4.8 and gpt-oss, whose selected states improved held-out correctness by 0.131 and 0.117, respectively. However, further distillation did not produce sustained increases: by round three, these gains had declined to 0.095 and approximately 0.010. These results show that improvement on acquisition tasks did not reliably predict the magnitude or persistence of improvement on unseen tasks.

Distilled skills could also be reused across models, although their benefit depended on the donor--recipient pairing (Fig.~\ref{fig:distill}f). Donor libraries outperformed the recipient's own distilled library in four of six tested pairings, with correctness gains of 0.010--0.089. For GLM-4.7-flash, libraries distilled by either Claude Opus 4.8 or GPT-5.6-sol increased held-out correctness from 0.363 with its own library to 0.452. For Qwen3.6:35b, the Claude-derived library increased correctness from 0.482 to 0.548, whereas the GPT-derived library yielded 0.476. Distilled skill libraries supported cross-model reuse, with variable effects on held-out performance.

We also compared the selected distilled libraries with the expert-written library on the same held-out tasks (Fig.~\ref{fig:distill}e). Distilled libraries did not uniformly outperform the expert-written library. For gpt-oss, correctness was 0.345 with its own distilled library and 0.405 with the expert-written library.

\subsection{Reproducible numerical results do not ensure accurate scientific judgments}\label{sec:res-scieval}

\begin{figure}[t!]
\centering
\includegraphics[width=\linewidth]{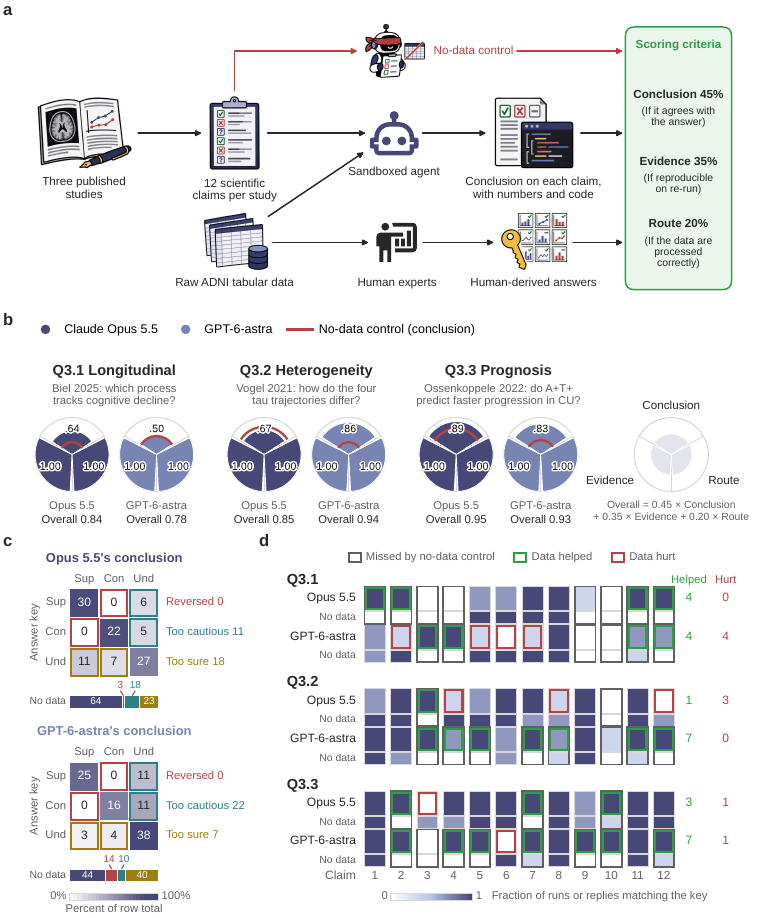}
\caption{\textbf{Evaluation of scientific judgment.} \textbf{a}, Overview of the claim-evaluation procedure. \textbf{b}, Performance of Claude Opus 5.5 and GPT-6-astra across the three scientific questions. Radial plots show conclusion agreement, evidence reproducibility and analytical-route scores. The overall score is the weighted sum of the three components. \textbf{c}, Confusion matrices comparing agent conclusions with expert-derived reference answers, pooled across the three questions and three runs. Outlined cells indicate incorrect conclusions, categorized by whether the agent was overly cautious or overly certain. Bars below the matrices show the corresponding distribution of no-data responses. \textbf{d}, Proportion of runs matching the reference answer for each claim, with data-enabled and no-data results shown separately.}\label{fig:adjudication}
\end{figure}

We evaluated whether agents could draw evidence-supported scientific conclusions from unfiltered Alzheimer's Disease Neuroimaging Initiative tables. The evaluation comprised 36 claims from three published studies of longitudinal cognitive change, tau-deposition trajectories and disease progression in cognitively unimpaired participants positive for amyloid and tau \cite{biel2025atn,vogel2021tau,ossenkoppele2022atpos}. Claude Opus 5.5 and GPT-6-astra each completed three independent runs per study, yielding 108 claim-level judgments per model. We compared their conclusions with expert-established reference answers and assessed whether their code reproduced the reported quantities and sample sizes and passed predefined checks for data-processing errors. The same models also answered the claims without data or tools, providing model-specific baselines (Fig.~\ref{fig:adjudication}a).

Numerical reproducibility consistently exceeded conclusion accuracy. All 18 data-enabled runs achieved the maximum evidence-reproducibility and analytical-route scores, regenerating the reported quantities and sample sizes and passing the predefined checks. Nevertheless, each model matched the reference conclusion in only 79 of 108 judgments (73.1\%), leaving 29 disagreements (26.9\%). Performance varied substantially across scientific questions (Fig.~\ref{fig:adjudication}b). Claude Opus 5.5 achieved conclusion accuracies of 63.9\% for longitudinal cognitive change, 66.7\% for tau trajectories and 88.9\% for disease progression; the corresponding values for GPT-6-astra were 50.0\%, 86.1\% and 83.3\%. Thus, successful re-execution and the absence of the checked processing errors left substantial uncertainty about whether agents would interpret their evidence correctly.

The models' identical aggregate accuracy concealed different judgment tendencies (Fig.~\ref{fig:adjudication}c). Neither model directly reversed a supported reference conclusion into contradicted, or vice versa. Instead, disagreements concerned whether the evidence justified a definite conclusion. Claude Opus 5.5 made 18 overly certain judgments, classifying undetermined claims as supported or contradicted, and 11 overly cautious judgments, withholding conclusions that the reference evidence supported. GPT-6-astra showed the opposite tendency, with 22 overly cautious and seven overly certain judgments. These differences indicate that aggregate accuracy alone obscures whether an agent tends to overstate evidence or overlook conclusions that it supports.

Access to data improved aggregate conclusion accuracy, but its benefit depended on both the model and the scientific question (Fig.~\ref{fig:adjudication}b,d). Across the three questions, accuracy increased from 59.3\% without data to 73.1\% with data for Claude Opus 5.5, and from 40.7\% to 73.1\% for GPT-6-astra. The tau-trajectory question showed particularly divergent effects: Claude Opus 5.5 declined from 75.0\% to 66.7\%, whereas GPT-6-astra improved from 33.3\% to 86.1\%. On this question, data changed Claude's majority response from incorrect to correct for one claim and from correct to incorrect for three; GPT improved on seven claims without corresponding losses. For longitudinal cognitive change, Claude improved from 33.3\% to 63.9\%, while GPT remained at 50.0\%, with improvements on four claims offset by losses on four others. Data therefore corrected some judgments while introducing new errors, and an unchanged aggregate score could conceal substantial changes in which claims were answered correctly.

\begin{figure}[t!]
\centering
\includegraphics[width=\linewidth]{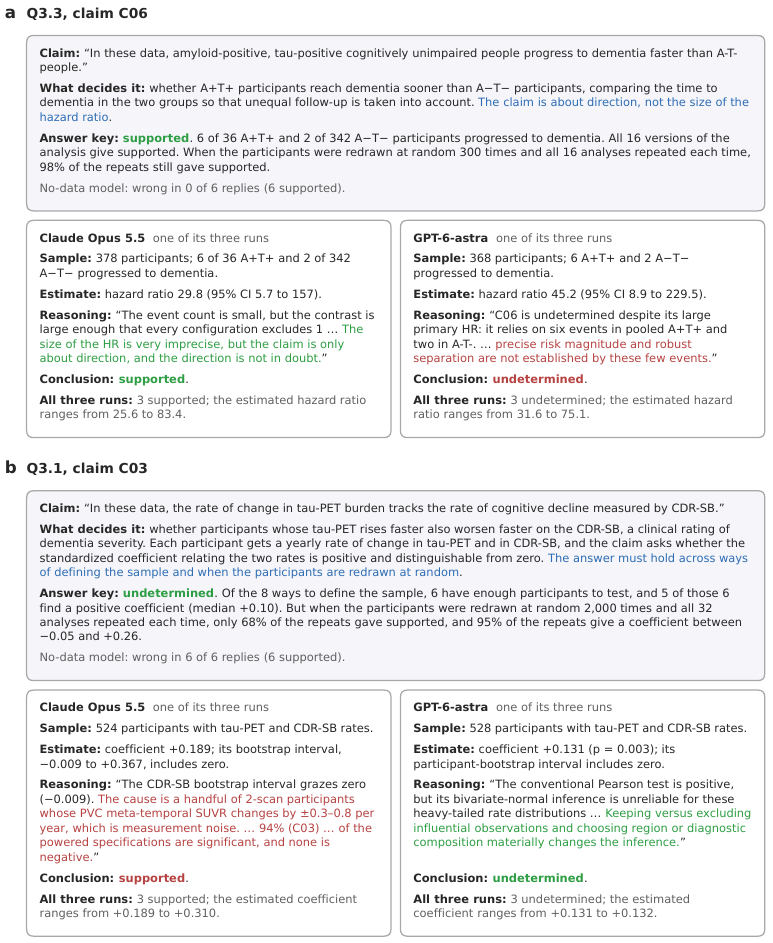}
\caption{\textbf{Examples of differences in scientific judgment between agents.} 
\textbf{a}, Evaluation of a claim concerning dementia progression in amyloid- and tau-positive cognitively unimpaired participants. \textbf{b}, Evaluation of a claim concerning the association between longitudinal tau-PET changes and cognitive decline.}\label{fig:examples}
\end{figure}

Two claims illustrate how these opposing errors arose when agents interpreted uncertainty (Fig.~\ref{fig:examples}). For the disease-progression claim, both agents found that six of 36 amyloid- and tau-positive participants progressed to dementia, compared with two of more than 340 negative participants. The reference conclusion was supported under all 16 analytical specifications and in 98\% of 300 participant-resampling repetitions. Claude Opus 5.5 agreed in all three runs, whereas GPT-6-astra classified the claim as undetermined in all three, emphasizing the small number of events even though its own 95\% confidence interval for the hazard ratio lay entirely above one. Conversely, a longitudinal association was reproduced in only 68\% of 2,000 participant-resampling repetitions, yielding an undetermined reference conclusion. Claude nevertheless classified it as supported in all three runs. In one run, it favored the finding that 94\% of its analytical variants were significant over its own bootstrap interval, which included zero; those variants reused the same participants. GPT's bootstrap interval also included zero, and it classified the claim as undetermined in all three runs. Together, these cases locate the judgment errors in how agents weighed evidence and uncertainty after obtaining reproducible numerical results.

% Figure 5, built by figures/src/fig5_examples.py. Every value and quote is read
% from the Q3.1/Q3.2 answer keys, the scored runs' verdicts.json, reports and
% code, the task instruction and the no-data replies at render time.

% Figure 6 is retained unchanged for placement and discussion in the
% Discussion section. It is not cited by the four Results subsections above.
\begin{figure}[t!]
\centering
\includegraphics[width=\linewidth]{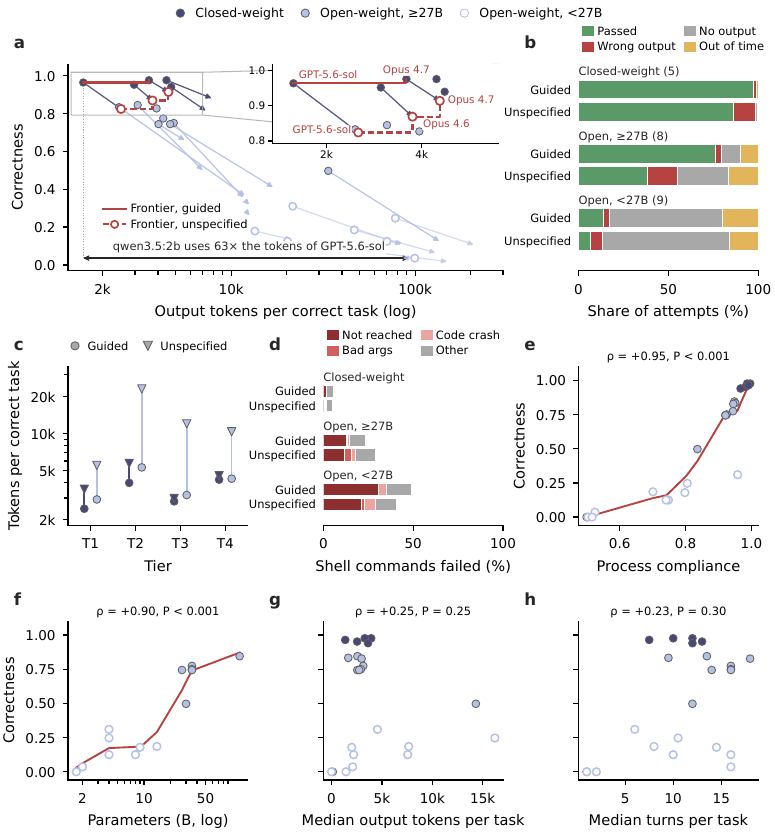}
\caption{\textbf{Computational efficiency and failure patterns across language models.} \textbf{a}, Correctness plotted against output tokens per correct task under protocol-guided and protocol-unspecified conditions. Arrows connect the two conditions for each model. Red lines indicate the Pareto frontiers, with the inset showing the highest-performing region. \textbf{b}, Distribution of task outcomes, including successful completion, incorrect outputs, missing outputs and timeouts, grouped by model class and instruction condition.
\textbf{c}, Median output tokens per correct task across the four task tiers. \textbf{d}, Proportion of failed shell commands by failure category, averaged across models within each class. \textbf{e}--\textbf{h},
Relationships between correctness and process compliance (e), model parameter count (f), median output tokens per task (g) and median turns per task (h). Curves in e and f are locally weighted regressions (LOWESS).}\label{fig:efficiency}
\end{figure}

\section{Discussion}\label{sec:discussion}

For neuroimaging researchers, the usefulness of autonomous agents depends on which analytical responsibilities they can reliably assume. NIBench shows that proficiency in executing a supplied protocol provides an incomplete account of this usefulness. Performance also depends on whether agents can independently construct workflows, reuse procedural experience and interpret the evidence they generate. These capabilities varied across the evaluated models and conditions, indicating that an agent's suitability should be assessed against its intended role in a study. This distinction connects benchmark performance to practical decisions about how much procedural guidance and scientific oversight an analysis requires.

Reusable procedural knowledge offers one route to reducing repeated instruction in neuroimaging analysis. The cross-model transfers show that experience acquired during task execution can be reused by another model through an external skill library. Such libraries could allow laboratories to inspect, revise and share analytical procedures as software and research practices evolve. However, improvements on acquisition tasks did not consistently extend to unseen tasks. A library's usefulness therefore needs to be established through independent evaluation, particularly when it is applied to a different analytical workflow or model.

The scientific-judgment evaluation identifies a further requirement for useful research assistance: agents need to assess what their numerical results support. Both evaluated models regenerated their reported evidence, yet each disagreed with the reference conclusion in 26.9\% of judgments. Their contrasting tendencies toward excessive certainty or caution show why successful computation alone provides limited assurance about scientific interpretation. This finding extends the concern raised by neuroimaging studies in which analytical choices affect conclusions \cite{botvinik2020variability,li2024variation}: evaluation also needs to examine how agents weigh uncertainty after completing an analysis. For research use, verification should therefore address both the reproducibility of the numerical evidence and the conclusion drawn from it.

The no-data comparison further clarifies what conclusion accuracy measures. Access to data improved aggregate performance, but its effects differed across scientific questions and sometimes introduced new errors. Correct answers obtained without analysis may reflect prior knowledge or inference from the claim wording; this control does not distinguish those possibilities or establish memorization. Comparing the two conditions helps identify the additional value of data analysis. Together with deterministic output checks and independent reference conclusions, this design separates successful task execution, reproducible evidence and agreement with the evaluated scientific conclusions.

These findings have several boundaries. Models were evaluated within a shared agent framework, so their performance may change with other configurations. The workflow suite uses fixed reference outputs, which can penalize scientifically defensible alternatives. Workflow comparisons and most skill-distillation comparisons lack repeated runs within each experimental condition, limiting conclusions about variability and the stability of small differences. Scientific judgment was assessed in two models using three studies from a single, extensively studied cohort. Moreover, the reference conclusions depend on claim-specific analyses and expert assessment of the resulting evidence. Broader evaluation across cohorts, research questions and expert-defined reference standards is needed to establish how well these judgment findings generalize.

NIBench provides a systematic reference for developing and evaluating autonomous agents for neuroimaging research. By distinguishing the capabilities involved in carrying out analyses and interpreting their results, it helps researchers relate measured performance to intended analytical responsibilities and helps developers identify where improvement is needed. Future advances can be assessed against both the reliability of workflow completion and the quality of the scientific judgments that follow.

\section{Methods}\label{sec:methods}

\subsection{Study design and agent systems}\label{sec:m-agent}

NIBench comprises three complementary evaluations: workflow execution and
reconstruction, acquisition and transfer of procedural expertise, and
scientific judgment. The workflow evaluation uses 56 neuroimaging tasks;
these tasks also provide the acquisition and held-out sets for skill
transfer. Scientific judgment is assessed using 36 claims from three
published Alzheimer's disease studies. The workflow evaluation includes 19 language models within a shared agent
framework: 11 closed-weight and eight open-weight models. Selected models
also undergo skill acquisition, transfer and scientific judgment evaluation. Repetition and analysis procedures are
described below.

The eight open-weight models in the workflow evaluation are
qwen3.5:122b-a10b, gpt-oss:latest, glm-4.7-flash, qwen3.5:35b,
qwen3.5:35b-a3b, qwen3.5:27b, qwen3.6:35b and qwen3:30b. The resource analysis in Fig.~\ref{fig:efficiency} includes five
closed-weight models (Claude Opus 4.8, Claude Opus 4.7, Claude Opus 4.6,
Claude Sonnet 4.6 and GPT-5.6-sol) and the same eight open-weight models.
It additionally includes nine smaller models to examine resource use and
failure patterns across a wider range of model sizes: gemma4:e4b,
qwen3:14b, qwen3.5:9b, qwen3:8b, qwen3:4b, qwen3.5:4b,
qwen3.5:2b, qwen3:1.7b and qwen3.5:0.8b.

All systems use the NeuroAgent framework \cite{neuroagent2026v1}, built on
the Claude Agent SDK, to inspect data, write code and invoke neuroimaging
software. The shared framework provides access to a library of 13 skills:
eight tool ecosystems, four imaging modalities and tool provisioning. The
tool ecosystems are AFNI \cite{cox1996afni}, ANTs \cite{avants2011ants},
dcm2niix \cite{li2016dcm2niix}, elastix \cite{klein2010elastix}, FreeSurfer
\cite{fischl2012freesurfer}, FSL \cite{jenkinson2012fsl}, MRtrix3
\cite{tournier2019mrtrix3} and PETPVC \cite{thomas2016petpvc}. Beyond task
instructions, these skills and tools are the only reusable external source of domain procedures supplied by the framework. Agents receive a catalogue and choose which skills to read.
The no-skills ablation removes this catalogue but tools are still provided.

Neuroimaging tools run in versioned Apptainer containers
\cite{kurtzer2017singularity}, including Neurodesk builds
\cite{renton2024neurodesk}. Reference answers and scoring code are hidden
from agents, and inputs are read-only. Runs have task-specific turn and
time limits. Agents are instructed not to access the network. After each run, we inspect
recorded commands and file edits for network activity.

\subsection{Workflow tasks and paired evaluation}\label{sec:m-suite}

The workflow suite covers structural, diffusion and functional MRI, PET and
multimodal analyses across seven tool ecosystems. It includes 16 atomic
operations, 23 multistep pipelines, five group analyses and 12 tasks with
adversarial inputs. The latter comprise six atomic, five pipeline-level
and one group-level task. Group analysis denotes a different analytical
scope rather than a uniform increase in computational difficulty. Clean
participants within adversarial tasks provide controls against general
failure.

Inputs come from OpenNeuro ds000114 \cite{gorgolewski2013ds000114}, IXI\cite{ixiDataset},
OpenNeuro ds002898\cite{jamadar2020simultaneous} and the Imaging Data Commons UPENN-GBM collection\cite{bakas2022university},
covering 35 participants. Atomic, pipeline and group tasks use two, three
and ten participants, respectively. Both instruction conditions use the
same input pool.

Each task specifies a scientific goal, required outputs and an analysis
protocol specifying tools, processing steps and parameter settings. Choices that can change the scientific answer, such as mask
definitions or positivity thresholds, remain part of the goal. The
protocol-guided condition provides all three components. The protocol-unspecified
condition removes only procedural guidance and instructs agents to use
standard tool defaults for implementation choices left unspecified by
the goal. Inputs, required outputs, references, tolerances and graders
remain identical across conditions. All 56 tasks have both variants.
This comparison measures the effect of withholding procedural guidance. Adversarial status is not disclosed through task titles or instruction
structure. Every task requires an informative report when an input is
unusable and prohibits fabricated deliverables. Expected values and
scoring rules are withheld.

\subsection{Workflow reference standards and scoring}\label{sec:m-grading}

Reference outputs are generated with recorded commands in fixed software
environments, checked against their graders and regenerated before the
suite is frozen. Inputs are checked for image and diffusion-gradient
problems and for answer leakage. Operations without reproducible
references are excluded or assigned documented tolerances based on measured
variability.

Graders compare delivered volumes, masks, tables, arrays,
scalars and reports with reference outputs. Checks include image shape and
affine agreement, mask overlap with a Dice threshold of 0.95, and numerical
agreement within specified absolute or relative tolerances. Reports must
describe failures informatively and cannot substitute fabricated artifacts.
No language model participates in correctness scoring.

For most tasks, correctness is the fraction of participants passing all
required checks. The five group tasks receive binary scores; the
adversarial group task averages table and reporting scores. A task passes
at correctness of at least 0.5, and suite correctness is the unweighted
mean of all 56 task scores. Timeouts receive zero and are reported
separately.

Process compliance is scored by assessing required skill access, forbidden
commands, preferred tools and adherence to the no-network policy, as
applicable to each task. Full correctness can therefore coexist with low
compliance.

\subsection{Procedural skill acquisition and transfer}\label{sec:m-sd}

Agents acquire procedural knowledge by distilling execution trajectories
into editable skill files without updating model weights. Each round runs
tasks using the previous skill state, scores the outputs, extracts
lessons from successful and failed trajectories, and consolidates the
lessons into a replacement library. The hand-written library remains
read-only, and the provisioning skill is frozen because it encodes
environment facts. The distiller sees agent-visible text, tool inputs and
abbreviated responses, but no reference artifacts, expected values,
tolerances or grader diagnostics. For failed runs, feedback states only
that the output did not meet the task contract; resulting lessons are
framed as checks rather than verified answers.

The 56 tasks are divided into two stratified halves of 28, matched by tier,
modality and tool ecosystem. Only the acquisition half contributes
trajectories; both halves are scored under each saved skill state. All
runs use protocol-unspecified instructions. Round~0 uses a minimal seed library
specifying container locations, which differs from the no-skills condition.
Up to three growth rounds are run. The selected state maximizes correctness
on the acquisition half, with earlier rounds winning ties. Held-out tasks
assess transfer beyond the acquisition set. Cross-model transfer holds
the recipient agent fixed while changing the library's author and scores
the held-out half. An append-only condition omits model-based
consolidation to assess the role of curation.

Leakage controls restrict distiller inputs and reject skills containing
benchmark-specific identifiers or private artifacts. A fixed probe model
receives the proposed library and task instruction without data or tools;
its attempted output is evaluated by the actual grader. The same probe
with an empty library provides a control. A task passed only with the
proposed library flags leakage, and that state is not scored.

\subsection{Scientific judgment evaluation}\label{sec:m-adjudication}

Three scientific questions use unfiltered Alzheimer's Disease Neuroimaging
Initiative (ADNI) tables \cite{petersen2010adni} and claims anchored in
studies of longitudinal cognitive change \cite{biel2025atn}, tau
trajectories \cite{vogel2021tau} and progression in amyloid- and
tau-positive cognitively unimpaired participants
\cite{ossenkoppele2022atpos}. Each question supplies twelve claims, the
source tables and a dictionary limited to those tables. Agents receive no
assembled analysis cohort and must select inclusion rules, measurements
and analytical models. For each claim, they report a supported,
contradicted or undetermined conclusion, the decisive quantity and sample
size, and executable code. These claims are separate from the workflow
task suite.

We establish reference answers through expert review of numerical results
from a reference analysis for each claim. The analysis reports the sample
size, effect estimates and uncertainty, together with sensitivity analyses
that examine relevant choices in participant inclusion, measurement and
statistical modelling. Experts consider these results in relation to the
claim's wording and the population it describes. They assess the direction
and magnitude of the effect, its uncertainty and how the findings change
under alternative analyses. Each claim receives a supported, contradicted
or undetermined label, accompanied by an explanation that cites the numerical
evidence. An undetermined label is used when the evidence is insufficient.

Each question combines conclusion agreement (weight 0.45), evidence
reproducibility (0.35) and analytical route (0.20). Agreement is the
fraction of twelve classifications matching the human-derived answers. Reproducibility
requires the delivered code to regenerate each claim's quantity and sample
size from the supplied inputs in a clean directory at the reported
precision. Route checks assess preidentified choices that can fail silently in the source tables.

Agents also classify the same claims without data, tools or files. We
compare these responses with the data-enabled runs to assess how access to
data changes conclusion accuracy. The no-data responses provide a baseline
that may reflect prior knowledge of published findings or inference from
the claim wording \cite{sainz2023contamination}.

We run one workflow sweep per model and instruction condition. Skill
acquisition and transfer comparisons use one run per experimental cell.
For scientific judgment, each model completes three data-enabled runs and
three valid no-data responses per question.

% \paragraph{Reporting summary}
% \TODO{Nature reporting-summary pointer (added at revision).}

% \bmhead{Data availability}
% The NIBench workflow task suite uses publicly accessible imaging datasets.
% The procedural expertise evaluation uses execution trajectories from this
% suite, while the scientific judgment evaluation uses ADNI tables subject to
% the ADNI data-use agreement and access requirements.

% \TODO{IXI (CC BY-SA), OpenNeuro ds000114/ds002898 (CC0), IDC UPENN-GBM
% (CC BY); NIBench workflow task manifest + rebuild scripts on GitHub/HF/Zenodo; ADNI
% derivatives DUA-gated with rebuild path.}

% \bmhead{Code availability}
% \TODO{repository URL(s); frozen commit; license. State that NIBench lives in
% the repository's \texttt{neurobench/} directory, its pre-publication name.}

\bmhead{Acknowledgements}
This work was supported by the National Institutes of Health (NIH) under grants R01EB022744, R01AG077578, R01AG064584, S10OD032285, U19AG078109, and P30AG066530.

Data used in preparation of this article were obtained from the Alzheimer’s Disease Neuroimaging Initiative (ADNI) database (adni.loni.usc.edu). As such, the investigators within the ADNI contributed to the design and implementation of ADNI and/or provided data but did not participate in analysis or writing of this report. A complete listing of ADNI investigators can be found at: http://adni.loni.usc.edu/wp-content/uploads/how\_to\_apply/\\ADNI\_Acknowledgement\_List.pdf

We thank the participants and investigators who contributed to the IXI dataset (https://brain-development.org/ixi-dataset/), OpenNeuro datasets ds000114 and ds002898, and the University of Pennsylvania glioblastoma (UPENN-GBM) collection. The OpenNeuro datasets were obtained from the OpenNeuro database, and the UPENN-GBM data were accessed through the National Cancer Institute Imaging Data Commons.

% Data used in preparation of this article were also obtained from the Dallas Lifespan Brain Study (DLBS) through OpenNeuro. We thank the DLBS participants and research team for making these data available.

% We thank the participants and investigators who contributed to the IXI dataset (https://brain-development.org/ixi-dataset/), OpenNeuro datasets ds000114 and ds002898, the Dallas Lifespan Brain Study (DLBS), and the University of Pennsylvania glioblastoma (UPENN-GBM) collection. The OpenNeuro datasets and DLBS data were obtained through OpenNeuro, and the UPENN-GBM data were accessed through the National Cancer Institute Imaging Data Commons.

% \bmhead{Author contributions}
% \TODO{CRediT.}

% \bmhead{Competing interests}
% The authors declare no competing interests.

% \bmhead{Use of AI-assisted technologies}
% \TODO{disclose AI-assisted writing/engineering per journal policy.}

\end{spacing}

\clearpage
\renewcommand{\refname}{References}
\begin{spacing}{0.9}
\bibliographystyle{style/naturemag}
\bibliography{refs}
\end{spacing}

% Extended Data tables removed 2026-09-26; sections/6_extended_data.tex is
% kept as the local record and is not compiled.
% Extended Data figures (NMI: up to 10 ED items, each cited in the main text).
% Inline while drafting; at submission each moves to its own upload.
% ED Fig. 1, built by figures/src/edfig_generations.py from data.py's GEN_*
% block (neurobench/scientific-questions/derive_q3_generations.py).
\clearpage
\section*{Extended Data}

\begin{figure}[h!]
\centering
\includegraphics[width=\linewidth]{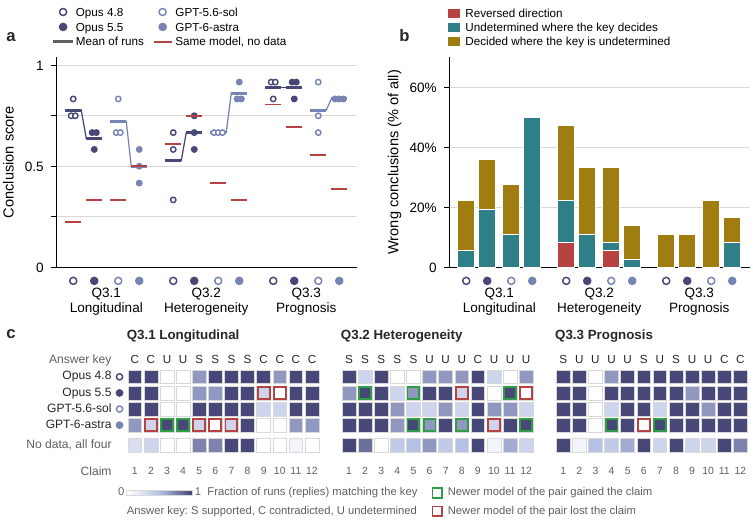}

\vspace{4pt}
% a minipage resets the figure's \centering, so the legend is justified
\begin{minipage}{\linewidth}\small\textbf{Extended Data Fig.~1 \textbar{} Comparison of scientific judgment across earlier and newer language models.} Scientific judgment was evaluated using Claude Opus 4.8, Claude Opus 5.5, GPT-5.6-sol and GPT-6-astra on the same three questions, with identical input data, instructions and reference conclusions. Each model completed three runs per question.
\textbf{a}, Conclusion accuracy for individual runs (circles) and the mean across three runs (horizontal bars). Open and filled circles distinguish earlier and newer models, respectively. Red lines indicate mean accuracy without access to data. \textbf{b}, Proportion of incorrect conclusions by error type, including reversed conclusions, undetermined responses to claims with definite reference conclusions, and definite responses to claims classified as undetermined. \textbf{c}, Claim-level conclusion accuracy across the four models and their no-data controls. Reference conclusions are indicated above each claim. Green outlines identify claims on which the newer model improved relative to its earlier counterpart, whereas red outlines indicate decreased accuracy.
\end{minipage}
\end{figure}

\end{document}